\documentclass[letterpaper,twocolumn,10pt]{article}

\usepackage{tikz}
\usepackage{amsmath}
\usepackage{cite}
\usepackage[normalem]{ulem}
\usepackage{mathptmx} 
\usepackage{fancyhdr}
\usepackage[normalem]{ulem}
\usepackage{upgreek} 
\usepackage{graphicx}
\usepackage{color, soul}
\usepackage[ruled, linesnumbered, vlined]{algorithm2e}
\usepackage{xspace}
\usepackage{multirow}
\usepackage{array}
\usepackage{comment}
\usepackage{listings}
\usepackage{microtype}
\usepackage{graphicx}
\usepackage{booktabs} 
\usepackage[normalem]{ulem}
\usepackage{lipsum} 
\usepackage{comment} 
\usepackage{xspace}
\usepackage{microtype}
\usepackage{graphicx}
\usepackage{scalerel}
\usepackage{booktabs}
\usepackage{tabularx}
\usepackage{enumitem}
\usepackage{balance} 
\usepackage{tikz} 
\usepackage{caption}
\usepackage{adjustbox}
\usepackage{subfig}
\usepackage{url} 
\usepackage{dblfloatfix}
\usepackage{enumitem}
\usepackage{subfig}
\usepackage{wrapfig}
\usepackage{xcolor}
\usepackage[capitalize,noabbrev,nameinlink]{cleveref}

\definecolor{codegreen}{rgb}{0,0.6,0}
\definecolor{codegray}{rgb}{0.5,0.5,0.5}
\definecolor{codepurple}{rgb}{0.58,0,0.82}
\definecolor{backcolour}{rgb}{1.0,1.0,1.0}

\lstdefinestyle{mystyle}{
    backgroundcolor=\color{backcolour},   
    commentstyle=\color{codegreen},
    keywordstyle=\color{magenta},
    numberstyle=\tiny\color{codegray},
    stringstyle=\color{codepurple},
    basicstyle=\ttfamily\footnotesize,
    breakatwhitespace=false,         
    breaklines=true,                 
    captionpos=b,                    
    keepspaces=true,                 
    numbers=left,                    
    numbersep=5pt,                  
    showspaces=false,                
    showstringspaces=false,
    showtabs=false,                  
    tabsize=2
}

\crefformat{section}{sec. #2§#1#3}
\crefformat{subsection}{#2\S#1#3}
\crefformat{subsubsection}{#2\S#1#3}
\Crefname{figure}{Figure }{Figs.}
\Crefname{equation}{Eqn.}{Eqns.}
\Crefname{section}{\S}{\S}

\newcommand{\TheName}{{Geode}\xspace} 

\usepackage{filecontents}

\newcommand{\authornote}[1]{%
  {\let\and\relax
  \renewcommand{\and}{\\\hspace{1em}}%
  \begin{tabular}[t]{c}#1\end{tabular}}}
  
\begin{document}

\date{}
\title{\Large \bf Geode: A Zero-shot Geospatial Question-Answering Agent with Explicit Reasoning and Precise Spatio-Temporal Retrieval}

\author{
\begin{tabular}{c}
{\rm Devashish V. Gupta} \hspace{2em} {\rm Azeez S. Ishaqui} \hspace{2em} {\rm Divya Kiran Kadiyala} \\[2ex]
Georgia Institute of Technology, Atlanta
\end{tabular}
}

\maketitle

\begin{abstract}

Large language models (LLMs) have shown promising results in learning and contextualizing information from different forms of data.
Recent advancements in foundational models, particularly those employing self-attention mechanisms, have significantly enhanced our ability to comprehend the semantics of diverse data types.
One such area that could highly benefit from multi-modality is in understanding geospatial data, which inherently has multiple modalities.
However, current Natural Language Processing (NLP) mechanisms struggle to effectively address geospatial queries.
Existing pre-trained LLMs are inadequately equipped to meet the unique demands of geospatial data, lacking the ability to retrieve precise spatio-temporal data in real-time, thus leading to significantly reduced accuracy in answering complex geospatial queries.
To address these limitations, we introduce Geode—a pioneering system designed to tackle zero-shot geospatial question-answering tasks with high precision using spatio-temporal data retrieval. Our approach represents a significant improvement in addressing the limitations of current LLM models, demonstrating remarkable improvement in geospatial question-answering abilities compared to existing state-of-the-art pre-trained models.

\end{abstract}

\section{Introduction}
\label{sec:intro}
Advancements in Large Language Models (LLMs) have ushered a new age of unprecedented improvement in text analysis, classification, and completion.
Especially with the self-attention mechanisms \cite{vaswani:attention}, parsing and contextualization of text based prompts has led to the rise of automated chat-bots and Question-Answering (QA) agents, such as chatGPT which can answer the user questions to a high degree of precision.
In addition to text based capabilities, LLMs are advancing to interact and learn from other forms of data such as images, video, and audio paving the way for Large Multi-modal Models (LMMs) such as Gemini \cite{anil:gemini}.\\

Owing to the recent and rapid development of Large Language Models (LLMs), developing a Large Geospatial Model (LGM) has become an important and tractable goal. These models leverage spatial information, integrating various data types such as satellite imagery, GIS (Geographic Information Systems) data, and environmental readings. The end goal using this data is to understand and predict patterns, trends, and relationships within geographical contexts.\\

IBM and NASA recently released a geospatial foundation model called \textit{Prithvi} \cite{jakubik:prithvi} that can achieve zero-shot multi-temporal cloud gap imputation, flood mapping, and wildfire scar segmentation. 
Researchers have also explored contrastive spatial pretraining paradigms like CLIP \cite{radford:clip}, for geospatial visual representations with promising results. However, due to a variety of modalities with different semantics and evolution dynamics, developing unified representations of geospatial data has been a challenge computationally and logistically. Additionally, language-guided interaction with geospatial data and spatio-temporal reasoning is a relatively unexplored area.\\

Although, pre-trained LLMs are repositories of extensive geospatial knowledge, yet unlocking this wealth presents a nuanced technical challenge. While conventional methodologies rely on leveraging geographic coordinates, such as latitude and longitude, to pinpoint specific locations, the crux lies in the LLMs' capacity to comprehensively grasp and interpret these numerical representations within real-world contexts. Furthermore, this challenge escalates when queries necessitate accessing real-time or spatio-temporal datasets, such as meteorological records, which may surpass the LLM's capability to interact with and extract requisite information.\\
 
Therefore, it is essential to develop a system that can integrate geospatial data and effectively address user inquiries demanding intricate coordination between the LLM and multiple data sources across diverse geospatial modalities. Moreover, this system must also have the ability to retrieve spatio-temporal data accurately from necessary sources, enabling the sophisticated analysis and execution of complex geospatial tasks.
In this context, we propose \TheName--a methodology and LLM system designed to perform zero-shot Question-Answering (QA) with explicit reasoning capabilities for complex geospatial queries, coupled with precise spatio-temporal data retrieval from open-source geospatial data sources.\\

\noindent In summary, we make the following contributions: %
\begin{itemize}[leftmargin=*]
    \item Build a multimodal, zero-shot capable, proof-of-concept Question-Answering agent (QA) with precise temporal and spatial retrieval, using open-source geospatial data.

    \item Develop various functional, model, and database experts for LLM to interact with diverse geospatial data modalities. 
    
    \item Implement the methodology with a streamlined tool chain and showcase its ability in reasoning over complex user queries to perform spatio-temporal geospatial tasks.
\end{itemize}

\noindent In this work we begin with a limited set of diverse geospatial modalities like topography and meteorological data to demonstrate the capabilities of our methodology. \\

\section{Background and Related work}
\label{sec:motivation}
Geographical Information Systems (GIS) integrate and analyze diverse geospatial data to generate digital thematic maps through computational techniques \cite{Kolios:gis}. 
GIS leverages fundamental principles of geography, cartography, and \TheName, empowering end-users to formulate queries, analyze spatial information, visualize data in maps, and present the final results as detailed thematic digital maps (e.g., Clarke \cite{clarke:GIS}; Maliene et al. \cite{Maliene:GIS}). 
The data handled by GIS systems encompasses a vast array of modalities, including vector and raster data formats, geodatabases, hyperspectral and multispectral data, and unstructured data. 
Notable examples of such systems include GeoSpark \cite{Yu:GeoSpark}, geoMesa \cite{Hughes:GeoMesa, Hulbert:GeoMesa}, GeoTrellis \cite{geotrellis:gis}, and RasterFrames \cite{rasterframes:gis}. 
These systems are typically employed to perform complex distributed processing on massive volumes of geospatial data across large clusters of high-performance computing systems, providing mission-critical insights for environmental protection, climate prediction, commerce, defense, and numerous other domains. 
However, these systems lack the ability to perform Natural Language Processing (NLP) tasks to address geospatial queries effectively. \\

Prior works in literature have focused on developing an LLM based interface to perform Natural Language Processing (NLP) tasks on Geospatial data. Mai et al. \cite{Mai:Sphere2Vec} showcased the practical applications of large language models (LLMs) in the geospatial domain which include tasks like recognizing fine-grained addresses, forecasting time-series data related to dementia records, and predicting urban functions. Zhang et al. \cite{zhang:geogpt} uses GeoGPT, an autonomous AI tool built upon GPT-3.5, designed to autonomously collect, process, and analyze geospatial data using only natural language instructions. However, both these works primarily utilized pre-trained LLMs without exploring the potential of fine-tuning these models to create a specialized foundation model tailored for geospatial applications. \\

In a separate study, Deng et al. (2023) developed K2, a language model specifically fine-tuned on a corpus of geoscience texts. This specialized model demonstrated remarkable performance on various natural language processing (NLP) tasks within the geoscience domain. However, the capabilities of K2 are still confined to common NLP tasks, such as question answering, text summarization, and text classification, limiting its applicability. \\

In contrast to the above works, a more recent approach taken in GeoLLM by Manvi et al. \cite{manvi:geollm}, employs an innovative method that can efficiently extract the vast trove of geospatial knowledge inherently embedded within LLMs by fine-tuning the LLMs on prompts carefully crafted with auxiliary map data obtained from open-source geospatial data repositories.
Furthermore, by fine-tuning multiple LLMs, the authors of GeoLLM aim to quantify and rigorously assess the extent of geospatial knowledge encapsulated within these models. 
Additionally, they seek to evaluate the scalability assessment of this knowledge for a wide range of practical geospatial tasks, including population density prediction. \\

Despite the fine-tuning with auxiliary map data, GeoLLM encounters limitations in tasks necessitating the retrieval of spatio-temporal data across varied modalities, sourced from  both real-time and offline sources. For instance, answering a question related to the air quality in a specific geographical location requires an ability to access and extract the real-time information, which potentially limits the ability of LLM to reason about complex queries on geospatial data.
Moreover, the zero-shot reasoning and complex mathematical computations (such as finding standard deviation) of LLMs are poor despite providing extensive prompting. 

In response, with \TheName, we aim to bridge the shortcomings by providing the LLM the capability to interact and retrive real-time data from diverse data modalities to complete the required tasks. In addition, the LLM is capable of performing complex reasoning and provide a more accurate answers to the geospatial queries.

\section{Methodology}
\label{sec:methodology}


\TheName is a system designed to answer geospatial user queries that may involve multiple modalities and require complex reasoning. We identified a set of key insights that informed the design of \TheName as a system. We observe that the zero-shot reasoning and mathematical computation capabilities of LLMs are poor \cite{peng:limitations, Zhuang:ToolQA, wu:Conic10k}. This can be improved to some extent with special prompting techniques like few-shot, chain-of-thought \cite{wei:chainOfThought} and tree-of-thought prompting \cite{Yao:TreeOfThought}. This is the reason why we opted to leverage explicit geospatial experts to compute answers to narrow subsets of the geospatial query space. This also leads to high degree of compositionality and extensibility to the system, as with implementation of on only a few additional experts can enable the system to answer a whole new set of queries in query space. Additionally, since geospatial inference should not be only limited to retrieval, we unify ML inference, retrieval and explicit reasoning within \TheName.

\begin{figure}[t]
    \centering
    \includegraphics[width=.99\columnwidth]{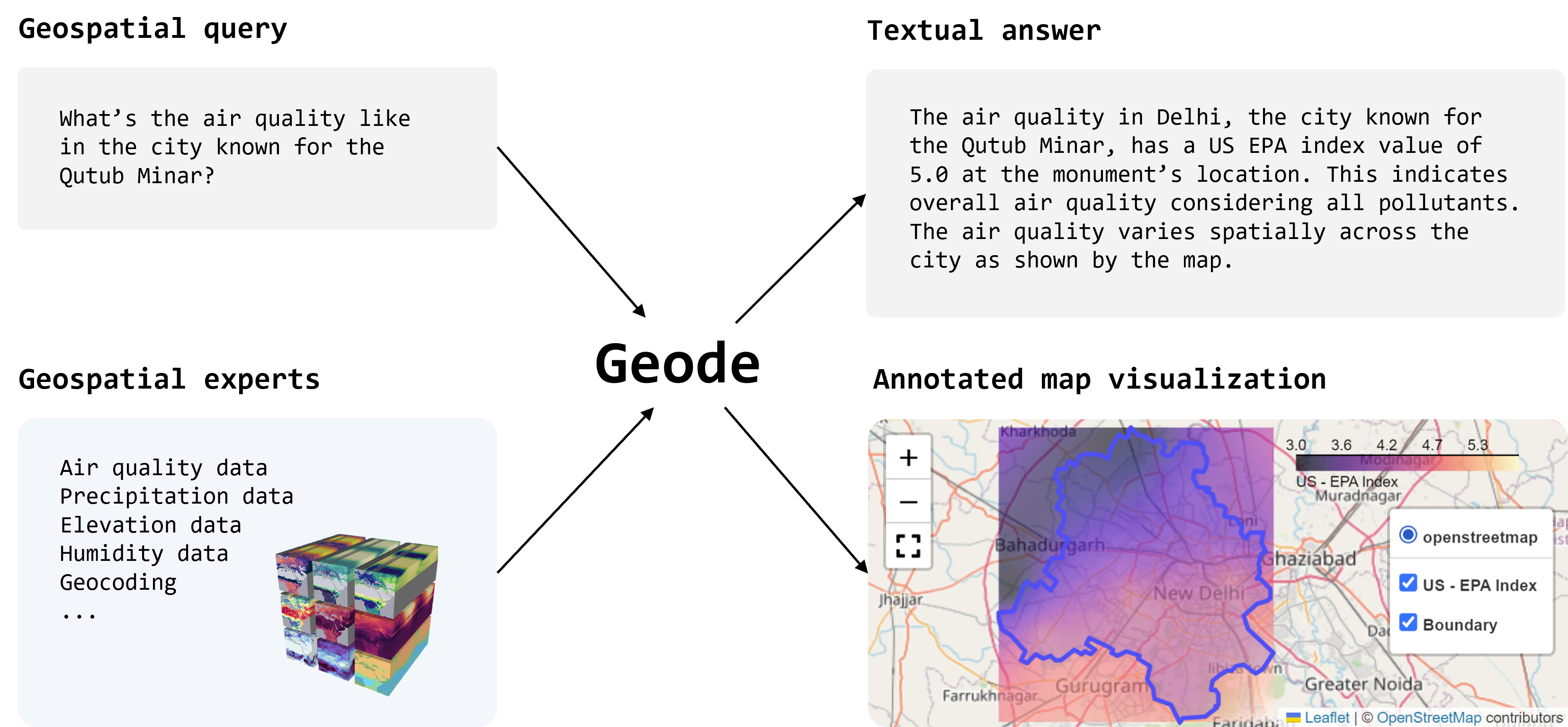}
    \caption{High level flow of proposed model for Geospatial data Question-Answering.}
    \label{fig:flow}
\end{figure}

\subsection{System Architecture}
\label{subsec:methdology:system-arch}
The architecture of \TheName is partly inspired by ViperGPT \cite{suris:viper}, which is a system designed to answer simple \textit{visual} queries on input images/videos based on a set of pretrained LLMs and VLMs as experts. ViperGPT relies on code generation to compose API calls that access the experts, followed by code execution to solve the given visual query. Its abilities include logical reasoning, ensuring consistency, mathematical operations, relations, negation etc. This paradigm of zero-shot code generation is a powerful way to leverage the knowledge of an LLM and augment it. However, ViperGPT can only perform visual inference and is not meant for geospatial question answering. Additionally, it only provides the final answer to the user along with code execution results to the user. This may be an output integer, float or a boolean value, which is not a great user experience. \\

\begin{figure*}[t]
    \centering
    \includegraphics[width=0.98\linewidth]{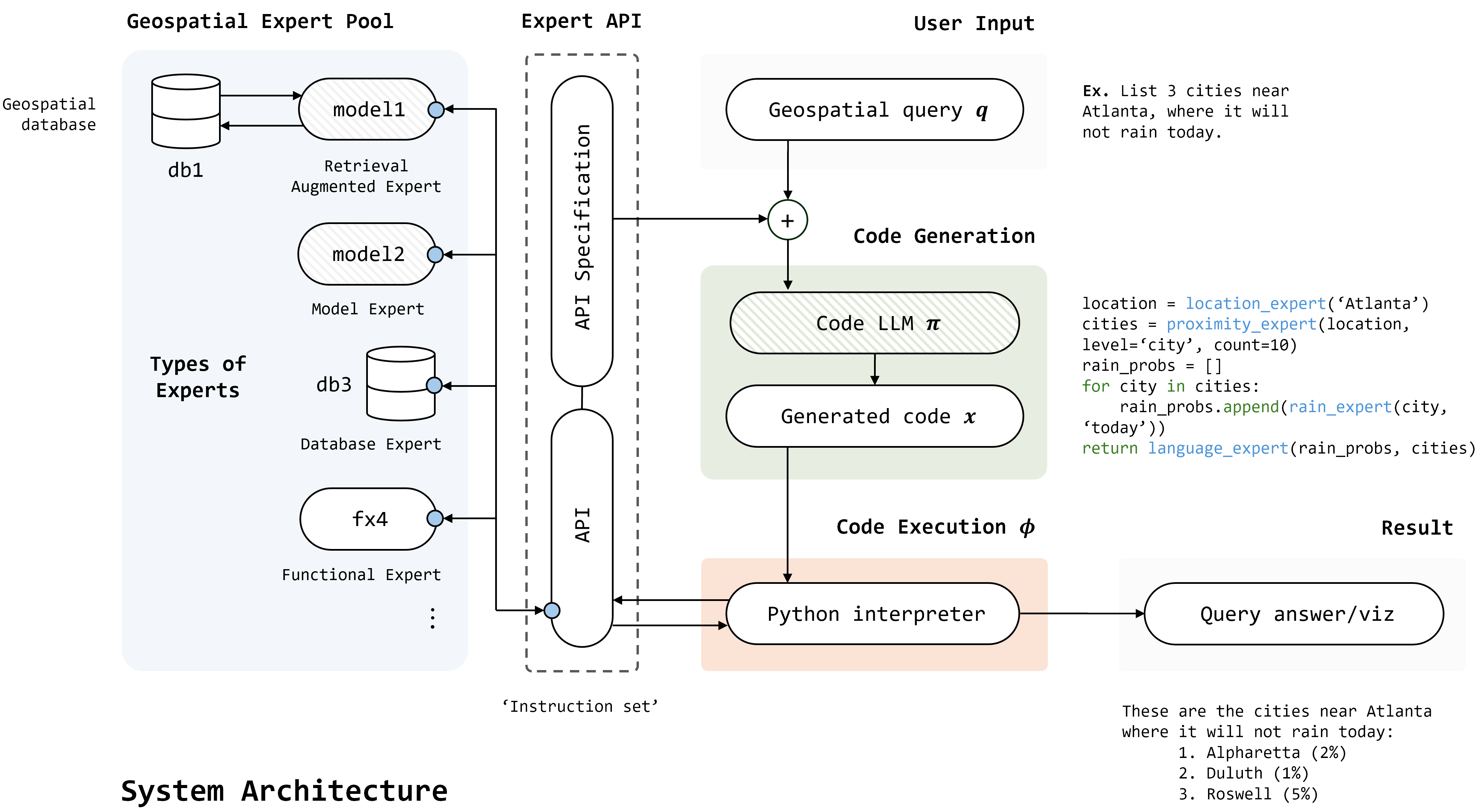}
    \caption{Geode System Architecture}
    \label{fig:sys-arch}
\end{figure*}

However, for geospatial inference, the results and computation often need to be visualized on a map and an textual explanation has to be given to the user to reveal the details on how the computation was done. This also improves the user experience for users who are not familiar with interpreting code. Keeping these aspects in mind, we built support for textual and map visualization for query computations within \TheName. Let us now look at the full architecture (\cref{fig:sys-arch}). \\

On a high-level, we have a geospatial expert pool that hosts all the geospatial experts which may be predictive ML models, geospatial databases and functional utilities. This expert pool is the heart of \TheName, where all the geospatial and modality specific computation is performed. The experts in the expert pool can be interacted with, using an expert API, which acts like an interface layer between the user and the expert pool. The expert API is decomposed into two parts namely the API specification and the full implementation. \\

In a similar fashion as ViperGPT, code generation is leveraged to understand and compose expert API calls in order to solve the user query. To provide the context and knowledge about the functionality available within the expert pool to the code generation model, we combine the API specification with the user query strategically to construct a prompt. The generated code is then executed in an execution environment finally generating a textual answer and salient artifacts for map visualization. The salience of artifacts is left up to the code generation LLM to determine. Let us examine each component of the system in a mode detailed fashion in the following section.

\section{Implementation}
\label{sec:implementation}


\subsubsection*{Geospatial expert pool}
As introduced previously, the geospatial expert pool is the heart of the system where all of the geospatial processing happens. Within this pool we built four major types of experts namely,

\begin{itemize}[topsep=0pt, leftmargin=*]
    \item \textit{Retrieval Augmented Experts}: These are language models, that have retrieval augmentation with geospatial data. This helps with reducing the occurrence of hallucinations and improve reliability of generated outputs.

    \item \textit{Model Experts}: These are ML models, specializing in a particular geospatial task like rain prediction, traffic flow prediction. These models do not require any retrieval augmentation since they are only good at one geospatial task.

    \item \textit{Database Experts}: These are geospatial databases with associated retrievers, access to data like realtime and historical weather, census, geography, nomenclature etc. This expert type allows for perfect retrieval, both spatial and temporal.

    \item \textit{Functional Experts}: These are utilities that may perform mathematical, analytical, geometric computations, which are not suitable for either ML inference or retrieval. For example, solving a differential equation, computing the vector intersection between multiple patches etc.
\end{itemize}

Here is the list of all the experts we implemented as part of the geospatial expert pool within Geode.

\begin{enumerate}[noitemsep, topsep=0pt, leftmargin=*]
    \item \texttt{point\_location\_expert}: Retrieves the point location or latitude and longitude of any place/address on Earth by its name.

    \item \texttt{patch\_location\_expert}: Retrieves the patch location of any place/address on Earth by its name, including its boundary polygon(s) and bounding box.

    \item \texttt{imputation\_expert}: Performs nearest neighbour data imputation on the raster data of any input \texttt{GeoPatch}. Useful for scenarios where data is not available for a particular location, but an estimate is needed for some downstream computation.

    \item \texttt{correlation\_expert}: Computes the cross-correlation between the raster data of two input \texttt{GeoPatch}'s

    \item \texttt{data\_to\_text\_expert}: Converts any python variable input into a human readable string format.

    \item \texttt{threshold\_expert}: Performs relative or absolute thresholding of the raster data within a \texttt{GeoPatch}, based on mode (greater/less).

    \item \texttt{intersection\_expert}: Performs either vector or raster intersection of data within the input geospatial patches. This also includes any data markers present within the \texttt{vector\_data} of the patches.

    \item \texttt{humidity\_expert}: Retrieves percent humidity values throughout any geographical patch as raster data, or at the central location of a patch based on mode (patch/point)
    
    \item \texttt{precipitation\_expert}: Retrieves precipitation values (in mm) throughout any geographical patch as raster data, or at the central location of a patch based on mode (patch/point)
    
    \item \texttt{temperature\_expert}: Retrieves temperature values (in Celcius) throughout any geographical patch as raster data, or at the central location of a patch based on mode.
    
    \item \texttt{air\_quality\_expert}: Retrieves a particular air quality parameter throughout a geographical patch as raster data, or at the central location of a patch based on mode. We support air quality parameters including, carbon monoxide, sulphur dioxide, nitrous oxide, ozone, PM2.5, PM10 and US EPA Index.
    
    \item \texttt{elevation\_expert}: Retrieves elevation values throughout a geographical patch as raster data, or at the central location of a patch based on mode.
    
    \item \texttt{elaborate\_expert}: Generates a elaborated textual answer based on the user query, computed final answer and any intermediate results as context.
    
    \item \texttt{patch\_visualization\_expert}: Visualize the vector and raster data in any patch and create an appropriate map visualization of the \texttt{GeoPatch}, no matter what it stores or represents.
    
\end{enumerate}

We implemented this basic set of experts as a proof-of-concept, while additional functionality may be introduced easily by implementing additional experts with no changes to the overall system. For geocoding capabilities, especially within point and patch location experts, we leveraged \texttt{Nominatim} from \texttt{geopy}. For the weather experts, we availed \texttt{WeatherAPI.com}  API for real-time weather data. For elevation data, we leveraged \texttt{OpenMeteo} elevation API. Many data modalities such as air quality and elevation are sparsely available globally, due to their dependence on meterological stations. To obtain a continuous estimate of the raster data parameter, we utilize bulk queries of the data at randomly sampled locations in a geospatial patch, followed by RBF kernel regression  for building a smooth and continuous estimate of the raster data.  \\

Having referenced the keyword \texttt{GeoPatch} multiple times, let us now look into what it is and what it represents. All of the experts mentioned above act upon a primary entity called a \texttt{GeoPatch}, which represents an arbitrary geospatial patch with corresponding vector and/or raster data. Shown below is the schema of the \texttt{GeoPatch} class.\\

\noindent\texttt{class GeoPatch():} Attributes
\begin{itemize}[topsep=0pt, leftmargin=*]
    \item \texttt{type: PatchType}
    Type of geographical patch based on the data it contains.

    \item \texttt{raster\_data: dict}
    Stores raster data and related information.
    \begin{itemize}[topsep=0pt, leftmargin=*]
        \item \texttt{name (str)}: Name of the raster data stored. (mandatory)
        \item \texttt{type (RasterType)}: Type of raster data stored, whether \texttt{RasterType.color}, \texttt{RasterType.non\_color}, or \texttt{RasterType.binary} (mandatory)
        \item \texttt{colormap (str)}: String representing a colormap name to best visualize the raster data. (optional)
        \item \texttt{data (np.ndarray)}: NumPy array containing the raster data. (mandatory)
    \end{itemize}

    \item \texttt{vector\_data: dict}
    Stores vector data and related information.
    \begin{itemize}[topsep=0pt, leftmargin=*]
        \item \texttt{location ([float, float])}: Coordinates of the location that the patch represents.
        \item \texttt{bbox (List[float])}: Bounding box coordinates of the boundary of the patch.
        \item \texttt{points (List[DataPoint])}: Data points corresponding to the patch, displayed on the map.
        \item \texttt{boundary (List[shapely.geometry.Polygon])}: Boundary polygons of the patch.
    \end{itemize}
    
\end{itemize}

\subsubsection*{User workflow}
\begin{figure*}[t]
    \centering
    \includegraphics[width=0.98\linewidth]{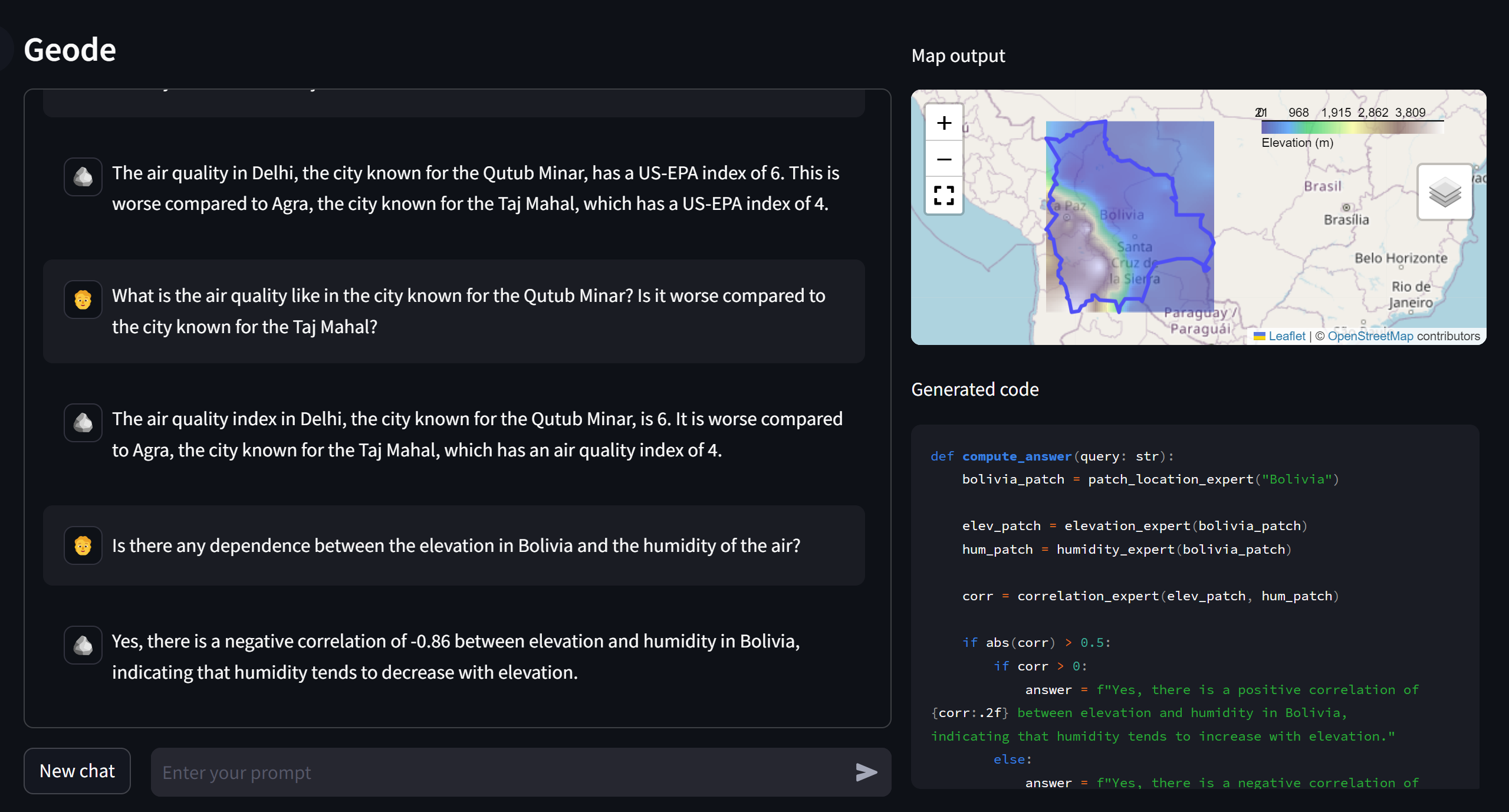}
    \caption{Geode User Interface}
    \label{fig:ui}
\end{figure*}

Having looked into the technical details of the geospatial expert pool, let us now examine how a user interacts with \TheName and what happens in the background. We used the \texttt{streamlit}  library to build an intuitive and easy to use front end for the app, which includes a chat container, a chat input area, a map visualization area and a generated code area. (Refer to \cref{fig:ui})\\

Whenever the user enters a geospatial query within the chat input, the query is fused into the base prompt, which contains the API specification. This gives us the full prompt, which includes the specifications for the \texttt{GeoPatch} and associated classes, specifications for the expert API calls, instructions for code generation and a code generation template (See Appendix \cref{sec:appendix}) that the model has to adhere to ensuring that any generated code can be executed successfully and the outputs can be visualized on the front-end, given the semantics of the code are correct. We built support for three code generation pipelines namely with, \texttt{OpenAI gpt-3.5-turbo-0125} \cite{gpt35turbo}, \texttt{Anthropic claude-3-opus} \cite{Claude} and local inference based on the quantized model \texttt{TheBloke/WizardCoder-Python-34B-V1.0-GPTQ} \cite{QGPT} from HuggingFace. For local inference, we implemented a custom backend using \texttt{Flask} \cite{Flask}. \\

The generated code is then validated with a set of safeguards that improve the probability of successful code execution, like removing any text prior to and following the code, escaping special characters in the strings that the code uses etc. Once the code is executed, the most salient \texttt{GeoPatch} is returned by the LLM for visualization in the map output area. For the interactive map visualization we employ \texttt{Folium} \cite{Folium} charts powered by \texttt{OpenStreetMaps} \cite{OpenStreetMap} and relied heavily on the \texttt{GeoJSON} \cite{GeoJSON} format. \\

Additionally, the generated code is shown to the user for improving the explainability of the system and allowing the user to examine the exact series of computations that led to the final answer. A well-formed textual answer is also provided to the user within the chat container, which previous systems like \texttt{ViperGPT} did not support for.

\section{Evaluation}
\label{sec:evaluation}

Our evaluation approach includes several different queries which test the capability of \TheName to perform QA which requires retrieval and augmentation of spatio-temporal data. While \TheName is able to deliver zero-shot responses for the easier queries, the more complex ones may require more prompting and elaboration on particular prompt aspects. We will be using the publicly available GIS datasets \cite{gis:gisgeography} to train and evaluate our proposed model.

\subsection{Evaluation Approach}
The evaluation process will involve a diverse range of user prompts, ranging from simple queries like \textit{"Where does it rain more, Atlanta or Chicago?"} to more complex tasks with detailed instructions, such as \textit{"Find the highest peak in Telengana"} These tasks necessitate the integration of various data modalities and the ability to associate events with specific times, setting this model apart from existing multi-modal models in its unique approach to handling intricate, time-sensitive geospatial queries. \\ 

The real-time retrieval capability of \TheName is a crucial aspect of its evaluation. By leveraging live data streams, the model can provide up-to-date and accurate responses to user queries. For instance, when a user asks about the current air quality index in a specific location, the model will retrieve the most recent data from relevant sources and incorporate it into its response. Similarly, when asked about current precipitation levels, the model will access live weather data to provide an accurate answer. This real-time retrieval feature ensures that the model's responses are not only relevant but also reflective of the current conditions at the time of the query. \\

In addition to real-time retrieval, \TheName model employs various expert components that specialize in extracting information from specific data modalities. These experts are designed to handle the unique characteristics and dynamics of each data type from raster data to JSON, enabling the model to provide comprehensive and accurate responses. For example, an expert focused on altitudes can identify hills and valleys across any region, while another expert specializing in climate data can predict weather patterns and trends. The evaluation process will assess the performance of these individual experts in their respective domains and their ability to contribute to the overall question-answering capability of \TheName model. \\

To measure the effectiveness of our \TheName model, we will employ a range of evaluation metrics that are embedded within the real-time retrieval and expert extraction processes.These metrics will be continuously monitored and analyzed throughout the evaluation period to provide a comprehensive assessment of the model's performance. Some of the key metrics we will consider include: 

\textbf{Response Latency}: Given the real-time nature of \TheName model, it is essential to evaluate its response latency. We will measure the time taken by the model to retrieve relevant data, process the query through the appropriate experts, and generate a response. Lower response latencies are desirable to ensure a smooth and efficient user experience. 

\textbf{Data Freshness}: as \TheName model relies on live data streams, we will assess the freshness of the data used in its responses. This metric will measure how up-to-date the information provided by the model is compared to the actual real-world conditions. Higher data freshness scores indicate that the model is effectively leveraging the most recent data available. 

\textbf{Expert Performance}: We will evaluate the performance of individual expert components in their respective domains. This evaluation will involve measuring the accuracy, completeness, and relevance of the information extracted by each expert. For example, we will assess the air quality index expert's ability to accurately report current air quality levels in green areas and the precipitation expert's accuracy in providing current rainfall data. 

\textbf{Code Completion}: To gauge the overall effectiveness of \TheName model, we must run generated code and measure the results of the visual output with the model's responses. This visual incorporation will provide valuable insights into the model's robustness, relevance, and ability to properly query. 

By incorporating these embedded evaluation metrics and continuously monitoring the model's performance in real-time with a visual code-completion mechanism, we can ensure a comprehensive and dynamic assessment of our \TheName model's capabilities. This evaluation approach not only validates the model's effectiveness in answering geospatial queries but also highlights its unique strengths in leveraging real-time data retrieval and expert extraction to provide accurate and up-to-date responses.

\begin{table}[]
\centering
\begin{tabular}{|m{8.0cm}|}
\hline
\textbf{Qutub Minar Example:} \\
\hline
\textbf{Geode} \\
\textit{Ask me anything Geospatial!} \\
\hline
\textbf{User Query} \\
What is the air quality like in the city known for the Qutub Minar? \\
\hline
\textbf{Geode Code Generation} \\
\begin{lstlisting}[language=Python,numbers=none]
def compute_answer(query: str):
    # Find the city known for the Qutub Minar
    qutub_minar_patch = patch_location_expert("Qutub Minar")
    city_patch = patch_location_expert("Delhi")
    # Get the air quality data for the city
    air_quality_patch = air_quality_expert(city_patch, parameter='us-epa-index', mode='patch')
    # Get a single point air quality value at the Qutub Minar location
    qutub_minar_aq = air_quality_expert(qutub_minar_patch, parameter='us-epa-index', mode='point')
    aq_value = qutub_minar_aq.vector_data['points'][0].data
    ...
    return text_answer, air_quality_patch
\end{lstlisting} \\
\hline
\vspace{2pt}
\textbf{Geode Map Output} \\
\vspace{1pt}
\includegraphics[scale=0.65]{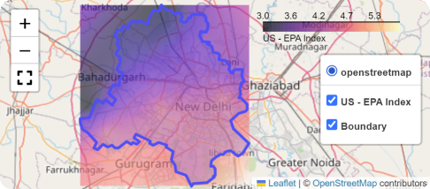} \\
\hline
\textbf{User Query} \\
What is the air quality like in the city known for the Qutub Minar? \\
\hline
\textbf{GPT-3.5} \\
The city known for the Qutub Minar is New Delhi, India. Unfortunately, New Delhi has faced significant air quality issues, especially during certain times of the year, such as in the winter months when pollution levels tend to spike due to various factors like vehicle emissions, industrial pollution, construction activities, and agricultural burning in neighboring states.
The air quality in New Delhi can vary widely from day to day and even hour to hour, but it has often been categorized as "very poor" or "severe" according to the Air Quality Index (AQI) scale, which measures the concentration of pollutants in the air. Measures are being taken by the government and various organizations to address this issue, including implementing stricter regulations on emissions and promoting the use of cleaner technologies. However, it remains a significant concern for the health and well-being of the city's residents. \\
\hline
\end{tabular}
\caption{Comparison of Geode response against GPT-3.5}
\label{table:result}
\end{table}

\subsection{Metrics of Success}
The primary metric of success for \TheName is its ability to accurately and efficiently answer user queries related to geospatial and temporal data for specified locations. We will evaluate the model's performance using the generated code and accompanying visual outputs, which show the completeness of its responses as shown in \cref{table:result}. Additionally, we will consider the response latency, data freshness, and expert performance metrics to ensure that the model provides timely, up-to-date, and accurate information. \\
 
In addition to these primary metrics, we will also perform ablation studies to evaluate the model's robustness and adaptability. These studies will involve testing the model's performance with different datasets, cross-domain knowledge questions, and varying levels of data availability. For example, a query within this could be something like \textit{"show me the correlation between precipitation and air quality in Bangladesh?"} By assessing the model's performance under diverse conditions with potentially limited data, we can identify its strengths, limitations, and  areas for enhancement. \\

Overall, the success of \TheName is determined by its ability to effectively integrate multiple data modalities, leverage real-time data retrieval and expert extraction, and provide accurate, relevant, and timely responses to user queries. By continuously monitoring and evaluating the model's performance using the aforementioned metrics, we aim to develop a robust and reliable geospatial question-answering agent that can serve as a valuable tool for a wide range of users and applications.

\section{Discussion}
\label{sec:discussion}

The development of \TheName is a significant step forward in the field of geospatial question-answering. By leveraging real-time data retrieval, expert extraction, and multi-modal data integration, our model demonstrates the potential to revolutionize how users interact with and derive insights from geospatial data.
One of the key strengths of our approach lies in its ability to provide accurate and up-to-date responses to user queries. The real-time retrieval capability ensures that the model always has access to live data, enabling it to generate responses that are up to date. This is particularly important in domains such as weather forecasting, air quality monitoring, and traffic management, where timely and accurate information is crucial for decision-making and public safety. \\

Moreover, the incorporation of expert components specializing in different data modalities allows \TheName to extract and analyze information from diverse sources effectively. By leveraging the unique characteristics and dynamics of each data type, these experts contribute to the model's ability to provide comprehensive and accurate responses. This modular approach also enables the model to adapt to new data sources and domains more easily, as additional experts can be developed and integrated as needed.
The evaluation of \TheName using a range of embedded metrics, including response latency, data freshness, expert performance, and code compilation provides a comprehensive assessment of its capabilities. The promising results obtained across these metrics demonstrate the model's effectiveness in answering geospatial queries accurately and efficiently. \\

However, it is important to acknowledge the limitations and challenges associated with our approach. The reliance on real-time data retrieval introduces potential issues related to data availability, quality, and consistency. Ensuring the reliability and robustness of data sources is crucial for maintaining the model's performance and credibility. Additionally, the integration of multiple data modalities and experts requires careful coordination and synchronization to avoid conflicts and inconsistencies in the generated responses. Code compilation serves to be a challenge as well, while the model may be successfully integrating everything internally, truly harnessing this code that runs successfully and displays meaningful output adds an additional layer of complexity.
Another challenge lies in the scalability and computational requirements of \TheName. As the volume and complexity of geospatial data continue to grow, the model's ability to process and analyze this data efficiently becomes increasingly important increasing the response latency with more complex queries. Optimizing the model's architecture, data storage, and retrieval mechanisms will be essential for ensuring its practicality and applicability in real-world scenarios. \\

Despite these challenges, the overall applications and impact of \TheName are far-ranging. From urban planning and environmental monitoring to emergency response and resource management, the ability to interact with geospatial data using natural language queries, particularly a powerful LLM, can empower a wide range of users and organizations. By democratizing access to geospatial insights, our model can facilitate data-driven decision-making, promote transparency, and foster collaboration across different domains. \\

Future work on \TheName shall focus on addressing the identified limitations and expanding its capabilities. This may involve exploring advanced data integration techniques, developing more efficient retrieval mechanisms, and develop further evaluation benchmarks to test the model's performance. We plan to also extend the model's application to new geospatial domains, like climate change analysis, biodiversity conservation, and public health data, to demonstrate its versatility and potential for impact.
\section{Conclusion}
\label{sec:conclusion}
\TheName is a powerful advancement in the field of geospatial question-answering. Through leveraging real-time data retrieval, expert composition, and multi-modal data integration, the system demonstrates the potential to revolutionize how users interact with and derive insights from geospatial data. While challenges related to data quality, scalability, and computational requirements remain, the promising evaluation results and potential applications of our model highlight its value and impact. As we continue to refine and expand the capabilities of \TheName, we envision a future where geospatial insights are accessible, actionable, and transformative for a wide range of users and domains.

\bibliographystyle{plain}
\bibliography{main}

\onecolumn
\section*{Appendix}
\label{sec:appendix}

\subsubsection*{Base prompt}

The base prompt consists of the \texttt{GeoPatch} and associated class specifications, expert API specifications, instructions on code generation and a template for the code to be generated so that the code execution results can be extracted and visualized.

\begin{lstlisting}[language=Python]
# API specification:

# helper classes
class RasterType(Enum):
    '''
    Enum representing the type of raster data stored in a patch.
    '''
    color = 0
    non_color = 1
    binary = 2

class DataPoint():
    '''
    Represents the data associated a particular point marker on the map.
    
    Attributes
    ----------
    point (shapely.geometry.Point): Stores the latitude and longitude of the data point in x and y.
    name (str): Name of the point, displayed within the tooltip on the map
    data (float): Any numerical value which can represent quanities like temperature, humidity etc.
    '''
    def __init__(self, x, y, name: str, data: float = None):
        self.point = Point(x, y)
        self.name = name
        self.data = data

# main class
class GeoPatch():
    '''
    Primary class representing a geospatial patch with vector/raster data.

    Attributes
    ----------
    raster_data: dict
        Stores raster data and related information.
        - 'name' (str): Name of the raster data stored. (mandatory)
        - 'type' (RasterType): Type of raster data stored, whether RasterType.color, RasterType.non_color, or RasterType.binary (mandatory)
        - 'colormap' (str): String representing a colormap name. (optional)
        - 'data' (np.ndarray): NumPy array containing the raster data. (mandatory)
    vector_data: dict
        Stores vector data and related information.
        - 'location' ([float, float]): Latitude and longitude of the location that the patch represents (mandatory).
        - 'bbox' (List[float]): Bounding box coordinates of the boundary of the patch [min_lat, max_lat, min_lon, max_lon] (mandatory).
        - 'points' (List[DataPoint]): Data points corresponding to the patch, displayed on the map (optional).
        - 'boundary' (List[shapely.geometry.Polygon]): Boundary polygon of the patch (mandatory).

    Methods
    -------
    get_raster_data() -> Dict
        Returns the raster_data present in the GeoPatch object
    set_raster_data(property: raster_data) -> None 
        Sets the given raster_data property into the GeoPatch object.
    set_raster_data_from_points(option_list: List[List[float]], name=None, type=None, colormap='gray') -> None
        sets the raster data for a list of list of points.
    get_vector_data() -> Dict
        gets the vector data from the object as Dictionary
    set_vector_data(property: vector_data) -> None
        sets the given vector data into the object
    get_boundary(boundary: List[Polygon]) -> None
        gets the boundary box as List of Polygon data
    get_bbox() -> List[float]
        gets the bounding box as List of floats
    get_location() -> List[float]
        Gets the location latitude and longitude as a list of floats

    '''
    def __init__(
            self, 
            raster_data: Union[Image.Image, np.ndarray] = None, 
            vector_data: Dict = None) -> None:
        '''
        Initialize a GeoPatch instance.
        
        Parameters
        ----------
        raster_data (Union[Image.Image, np.ndarray]): Raster data and related information (optional)
        vector_data (Dict): Vector data and related information (optional)
        '''
        self.raster_data = raster_data
        self.vector_data = vector_data

    def get_raster_data(self) -> Dict:
        '''
        Get the raster data stored in the GeoPatch.
        
        Returns
        -------
        Dict: Raster data and related information.
        ''' 
        if self.raster_data is not None:
            return self.raster_data

    def set_raster_data(self, raster_data: Dict) -> None:
        '''
        Set the raster data for the GeoPatch.
        
        Parameters
        ----------
        raster_data (Dict): Raster data and related information.
        '''
        self.raster_data = raster_data

    def set_raster_data_from_points(self, points: List[List[float]], name=None, type=None, colormap='gray') -> None:
        '''
        Sets the raster data across the patch from a list of points
        
        Parameters
        ----------
        points (List[List[float]]): List of points of the form [[lat0, lon0, value0], ...]
        name (str): Name of the raster data (optional)
        type (RasterType): Type of the raster data. Possible values: [RasterType.color, RasterType.non_color, RasterType.binary] (optional)
        colormap (str): Colormap for the raster data.
        '''

    def get_vector_data(self) -> Dict:
        '''
        Get the vector data stored in the GeoPatch, potentially containing 'points', 'boundary', 'location', 'bbox'.
        
        Returns
        -------
        Dict: Vector data and related information.
        '''
        return self.vector_data
    
    def set_vector_data(self, vector_data) -> None:
        '''
        Set the vector data for the GeoPatch. Mandatory keys must be present.
        
        Parameters
        ----------
        vector_data (Dict): Vector data and related information.
        '''
        self.vector_data = vector_data

    def get_boundary_polygons(self) -> List[Polygon]:
        '''
        Get a list of boundary polygons of the geographic location represented by the patch.
        
        Returns
        -------
        List[Polygon]: List of shapely.geometry.Polygon representing the boundary
        '''
        return self.vector_data['boundary']

    def get_area(self) -> float:
        '''
        Gets boundary area for the patch in million sq km.

        Returns
        -------
        float: Area of the patch in million sq km.
        '''

    def get_bbox(self) -> List[float]:
        '''
        Get the bounding box coordinates of the patch.
        
        Returns
        -------
        List[float]: Bounding box coordinates, format: [min_lat, max_lat, min_lon, max_lon]
        '''
        if 'bbox' in self.vector_data:
            return self.vector_data['bbox']

    def get_location(self) -> List[float]:
        '''
        Get the latitude and longitude of the patch.
        
        Returns
        -------
        List[float]: Latitude and longitude.
        '''
        if 'location' in self.vector_data:
            return self.vector_data['location']
        return None

    def get_data_points(self) -> List[DataPoint]:
        '''
        Get the data points associated with the locations within the patch.

        Returns
        -------
        List[DataPoint]: list of data points containing latitude, longitude, name and data.
        '''
    
    
# Here are all the geospatial experts you have access to as API calls:
def point_location_expert(name: str) -> GeoPatch:
    '''
    Finds the geographic location of any place on the map by its name.

    Parameters
    ----------
        name (str): Name of the place for which the geo-location has to be found.

    Returns
    -------
        GeoPatch: Geographical patch with the location and boundary of the place.
    '''

def patch_location_expert(name: str) -> GeoPatch:
    '''
    Finds the geographic location and boundary polygon of any place on the map by its name.

    Parameters
    ----------
        name (str): Name of the place for which the geo location has to be found

    Returns
    -------
        GeoPatch: GeoPatch containing the location and boundary path of the found place.
    '''

def imputation_expert(patch: GeoPatch) -> GeoPatch:
    '''
    Impute missing values in a patch using interpolation.

    Parameters
    ----------
        patch (GeoPatch): Input patch with missing values within patch.raster_data['data'] represented as NaN.

    Returns
    -------
        GeoPatch: Patch with missing values imputed.
    '''

def correlation_expert(patch1: GeoPatch, patch2: GeoPatch) -> float:
    '''
    Cross-correlate the raster data within two input patches.

    Parameters
    ----------
        patch1 (GeoPatch): First patch.
        patch2 (GeoPatch): Second patch.

    Returns
    -------
        float: Value of correlation between the raster data in patch1 and patch2
    '''

def data_to_text_expert(data: any) -> str:
    '''
    Computes the string representation for any input data.

    Parameters
    ----------
        data (any): Input data which is to be represented as a string.

    Returns
    -------
        str: String representation of the input data.
    '''

def threshold_expert(patch: GeoPatch, threshold: float, mode: str = 'greater', relative: bool = True) -> GeoPatch:
    '''
    Threshold the raster data within a GeoPatch by a percent or absolute threshold.

    Parameters
    ----------
        patch (GeoPatch): A GeoPatch whose raster data is to be thresholded.
        threshold (float): Value between 0.0 and 1.0, the percent threshold.
        mode (str): Possible values: ['greater', 'less']. Mode specifying the truth of a pixel value when compared to the threshold.
        relative (bool): Whether to use a percent or absolute threshold.

    Returns
    -------
        GeoPatch: Patch with thresholded raster data, RasterType of the GeoPatch.raster_data changes to RasterType.binary
    '''

def intersection_expert(patch1: GeoPatch, patch2: GeoPatch, mode: str = 'raster') -> GeoPatch:
    '''
    Perform intersection between the vector or raster data within two geographical patches. 
    If raster intersection is to be performed, both patches should have RasterType.binary and cover identical geographical regions.

    Parameters
    ----------
        patch1 (GeoPatch): First patch
        patch2 (GeoPatch): Second patch
        mode (str): Intersection mode, Possible values: ['vector', 'raster']

    Returns
    -------
        GeoPatch: with required intersection present within vector or raster data.
    '''

def humidity_expert(patch: GeoPatch, mode: str = 'patch') -> GeoPatch:
    '''
    Retrieves humidity (%) values throughout a geographical patch as raster data, or at the central location of a patch based on mode.

    Parameters
    ----------
        patch (GeoPatch): Geographical patch for which the humidity is to be retrieved.
        mode (str): Possible values: ['patch', 'point']. To specify whether to retrieve humidity data for the entire patch or a single point.

    Returns
    -------
        GeoPatch: patch with percent humidity as raster data, if mode == 'patch' 
            or else patch with patch.vector_data['points'][0].data as the humidity value, if mode == 'point'
    '''

def precipitation_expert(patch: GeoPatch, mode: str = 'patch') -> GeoPatch:
    '''
    Retrieves precipitation values in mm throughout a geographical patch as raster data, or at the central location of a patch based on mode.

    Parameters
    ----------
        patch (GeoPatch): Geographical patch for which the precipitation is to be retrieved.
        mode (str): Possible values: ['patch', 'point']. To specify whether to retrieve precipitation data for the entire patch or a single point.

    Returns
    -------
        GeoPatch: patch with precipitation (mm) as raster data, if mode == 'patch' 
            or else patch with patch.vector_data['points'][0].data as the precipitation value, if mode == 'point'
    '''

def temperature_expert(patch: GeoPatch, mode: str = 'patch') -> GeoPatch:
    '''
    Retrieves temperature values (Celcius) throughout a geographical patch as raster data, or at the central location of a patch based on mode.

    Parameters
    ----------
        patch (GeoPatch): Geographical patch for which the temperature is to be retrieved.
        mode (str): Possible values: ['patch', 'point']. To specify whether to retrieve temperature data for the entire patch or a single point.

    Returns
    -------
        GeoPatch: patch with temperature as raster data, if mode == 'patch' 
            or else patch with patch.vector_data['points'][0].data as the temperature value, if mode == 'point'
    '''

def air_quality_expert(patch: GeoPatch, parameter: str = 'pm2_5', mode: str = 'patch') -> GeoPatch:
    '''
    Retrieves a particular air quality parameter throughout a geographical patch as raster data, or at the central location of a patch based on mode.

    Parameters
    ----------
        patch (GeoPatch): Geographical patch for which the air quality index is to be evaluated. patch has latitude and longitude information.
        parameter (str): The air quality parameter to be retrieved. Possible values: ['co', 'no2', 'o3', 'so2', 'pm2_5', 'pm10', 'us-epa-index']
        mode (str): Possible values: ['patch', 'point']. To specify whether to retrieve air quality data for the entire patch or a single point.

    Returns
    -------
        GeoPatch: patch with the chosen air quality parameter plotted as raster data accessible as GeoPatch.raster_data['data'], if mode == 'patch'
            or else patch with patch.vector_data['points'][0].data as the air quality parameter value, if mode == 'point'
    '''
    param_info = {
        'co': ('Carbon Monoxide (ug/m3)', 'Greys'),
        'no2': ('Nitrogen dioxide (ug/m3)', 'Oranges'),
        'o3': ('Ozone (ug/m3)', 'Blues'),
        'so2': ('Sulfur Dioxide (ug/m3)', 'YlOrBr'),
        'pm2_5': ('PM2.5 (ug/m3)', 'magma'),
        'pm10': ('PM10 (ug/m3)', 'magma'),
        'us-epa-index': ('US - EPA Index', 'magma')
    }

def elevation_expert(patch: GeoPatch, mode: str = 'patch') -> GeoPatch:
    '''
    Retrieves elevation values throughout a geographical patch as raster data, or at the central location of a patch based on mode.

    Parameters
    ----------
        patch (GeoPatch): Geographical patch for which the elevation is to be retrieved.
        mode (str): Possible values: ['patch', 'point']. To specify whether to retrieve elevation data for the entire patch or a single point.

    Returns
    -------
        GeoPatch: patch with elevation (m) as raster data, if mode == 'patch'
            or else patch with patch.vector_data['points'][0].data as the temperature value, if mode == 'point'
    '''

# Instructions:
You are a powerful code generation model which can solve geospatial queries using the experts you have access to. 
Please write an implementation for a function 'compute_answer' using the expert API calls and classes you have access to, such that the answer to the query is obtained and returned:

# Query:
QUERY_TAG

# Your output should be exactly in this format and should not include any text before or after:
def compute_answer(query: str):
    # implementation using expert API calls here

    # returning a textual answer to the query and the most salient GeoPatch object to be used for visualization of the answer on a map
    return answer, patch 
result = compute_answer(query) # make sure you call the compute_answer function at the end and store the output in a variable called result



\end{lstlisting}

\end{document}